# Symbolic Probabilistic Inference with Evidence Potential

Kuo-Chu Chang and Robert Fung
Advanced Decision Systems
1500 Plymouth Street
Mountain View, California   94043-1230

## Abstract

Recent research on the Symbolic Probabilistic Inference (SPI) algorithm[2] has focused attention on the importance of resolving general queries in Bayesian networks. SPI applies the concept of dependency-directed backward search to probabilistic inference, and is incremental with respect to both queries and observations. In response to this research we have extended the evidence potential algorithm [3] with the same features. We call the extension symbolic evidence potential inference (SEPI). SEPI like SPI can handle generic queries and is incremental with respect to queries and observations. While in SPI, operations are done on a search tree constructed from the nodes of the original network, in SEPI, a clique-tree structure obtained from the evidence potential algorithm [3] is the basic framework for recursive query processing.

In this paper, we describe the systematic query and caching procedure of SEPI. SEPI begins with finding a clique tree from a Bayesian network — the standard procedure of the evidence potential algorithm. With the clique tree, various probability distributions are computed and stored in each clique. This is the "pre-processing" step of SEPI. Once this step is done, the query can then be computed. To process a query, a recursive process similar to the SPI algorithm is used. The queries are directed to the root clique and decomposed into queries for the clique's subtrees until a particular query can be answered at the clique at which it is directed. The algorithm and the computation are simple. The SEPI algorithm will be presented in this paper along with several examples.

## 1 Introduction

The Bayesian networks technology provides a representation language for uncertain beliefs and inference algorithms for drawing sound conclusions from such representations. Bayesian Network is a directed, acyclic graph in which the nodes represent random variables, and the arcs between the nodes represent possible probabilistic dependence between the variables. The success of the representation is mainly due to the development of many probabilistic inference algorithms [3, 4, 5, 6]. While most of the algorithms can efficiently perform simple queries such as the marginal probability of each node given evidence, they have not efficiently addressed the problem of more general queries such as joint or conditional probabilities of any combination of nodes.

The recent work of Symbolic Probabilistic Inference (SPI) [1, 2] has made a significant step in this direction. SPI is a goal-driven method which is incremental with respect to both queries and observations. In response to this research we have extended the evidence potential (EP) algorithm [3] with the same features. We call the extension symbolic evidence potential inference (SEPI). Unlike traditional Bayesian Net inferencing algorithms, both SPI and SEPI are goal directed, performing only those calculations that are required to respond to queries. While in SPI, operations are done on a search tree constructed from the original network, in SEPI, a clique-tree structure obtained from the EP algorithm is the basic framework for recursive query processing.

In SEPI, the EP algorithm [3] is used as the "pre-processing" step in which various probabilities such as "set-chain" conditional [3] and marginal probabilities of each clique are computed based on the clique tree. The second step in SEPI is to process the query with a recursive mechanism similar to the SPI algorithm. A query is directed to the root clique and decomposed into queries for the clique's subtrees. This recursive process continues until a particular query can be answered at the clique at which it is directed. The answer is then computed and returned to the next higher level in the clique tree. Once a clique has responses from all of its subtrees it can compute its own response to its predecessor clique. This process terminates when the root clique processes all the responses from its subtrees.

With similar mechanisms for caching and incorporating evidence as in SPI, the calculation in SEPI is also incremental with respect to both query and evidences. However, since all the necessary probability distributions are stored in the "pre-processing" step, the SEPI algorithm is more efficient.

The paper is organized as follows. Section 2 briefly describes the EP algorithm which includes the construction of the clique tree. Section 3 describes the SEPI algorithm. A systematic recursive query and caching procedure will be presented. Some illustrative examples are given in Section 4, followed by the concluding remarks in Section 5.

## 2  Evidence Potential Algorithm

In this section, we will briefly review the evidence potential (EP) algorithm [3]. The algorithm first organizes the original network into clique tree, where each clique is a group of nodes not necessary mutually exclusive. It then performs inference by passing messages between cliques in a similar way to the distributed algorithm [4].

The first part of the algorithm is to form a clique tree. This part consists of five steps

1. Marry Parents: link predecessors of a node together
2. Remove Arc Directions: remove directions of all arcs
3. Fill in: generate new arcs between nodes whenever necessary to form a "perfect" graph
4. Find Cliques: form node clusters/cliques
5. Order Cliques, and Find Residuals and Separators: form cluster tree

After the clique tree is formed, the second part of the algorithm is to calculate the marginal probability of each node. Before this can be done, the "evidence potential" and "separator potential" likelihoods [3] are calculated for each cluster.

The second part of the algorithm consists of the following:

1. Calculating Evidence Potentials and Separator potentials: they are calculated from the prior node conditionals in each clique.
2. Calculate Set-Chain Conditionals: namely, the conditional probability of the residuals given the separators of each clique.
3. Calculate Joint Probability for each clique: from joint, we then can calculate individual node posteriors (marginal) probability.

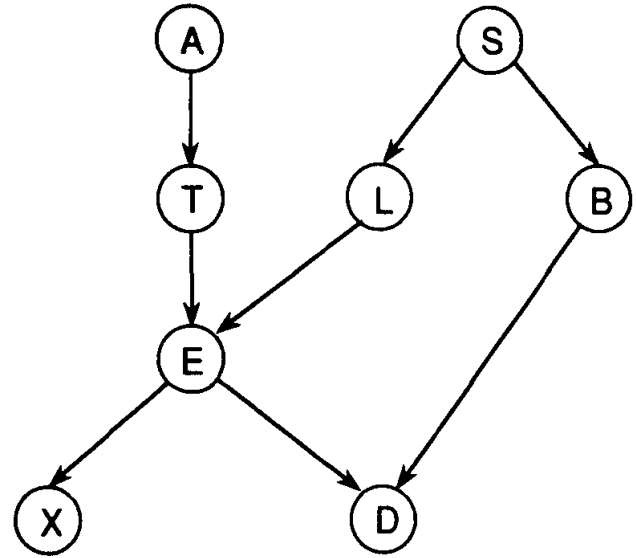

Figure 1: A Example Network

To illustrate this algorithm, the example given in [3] is shown in Figure 1. The corresponding clique tree and the set-chain conditional of each clique is shown in Figure 2. It is clear that the joint probability of the whole network can be obtained by multiplying all the set-chain conditionals together with the marginal probabilities of all the root cliques. Any query can then be obtained from the joint probability. The basic idea of EP algorithm is to decompose and factor the original formulae so that only minimum operations are required to answer the queries.

## 3  Symbolic Inference with Evidence Potential

The procedure described in the previous section can be considered as the pre-processing step for the generic query algorithm to be described. We call this new algorithm symbolic evidence potential inference (SEPI). In this algorithm, the goal is to calculate the results of arbitrary queries. The idea is to derive an efficient inference algorithm which takes advantage of the clique-tree structure of the EP algorithm.

The SEPI algorithm consists of several major processing steps. The first step is to organize the nodes of a Bayesian network into a clique tree structure and calculate and store the various probability functions as described in the previous section (e.g., set-chain conditional and joint probability distribution). In the second step, queries from the user are directed to the root clique of the tree. The query is decomposed into queries for the clique's subtrees. This recursive procedure continues until a particular query can be answered.

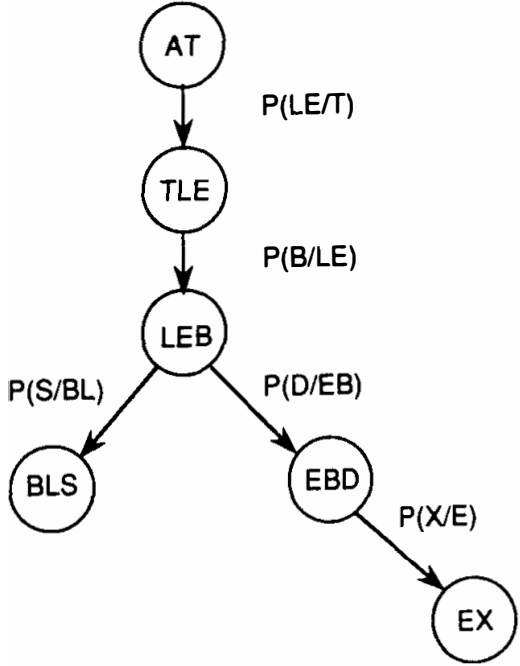

Figure 2: Cluster Tree and Set-Chain Conditional

The general format for a query received by SEPI is as a conditional probability, namely, $P\{X|Y\}$, where $X$ and $Y$ are sets of nodes in the network. This query is first transformed into joint distribution format $P(Z) \doteq P\{X,Y\}$ and directed to the root clique. In order to answer the query, it would be sufficient to calculate $P(Z \setminus C_0 | C_0)$, where $C_0$ is the root clique. This is because we can calculate the query by

$$P(Z) = \sum_{C_0 \setminus Z} P(Z \setminus C_0 | C_0) P(C_0) \quad (1)$$

where the prior probability $P(C_0)$ is available at the clique $C_0$. According to the EP algorithm, the clique tree is organized in a way that the separators are the overlapping nodes between the successor and predecessor cliques. Denote the separators between the root clique and the child clique $C_i$ as $S_i$, then

$$P(Z \setminus C_0 | C_0) = P(Z \setminus C_0 | S_i) \quad (2)$$

Define $T(C_i)$ as all the nodes in the subtree rooted from $C_i$, then the new request to be sent to each child $C_i$ is

$$P((Z \setminus C_0) \cap T(C_i) | S_i) \quad (3)$$

Note that if the successor clique has nothing to do with the query, i.e., $(Z \setminus C_0) \cap T(C_i) = \emptyset$, then no query will be sent to that clique.

At the clique $C_i$, when the request arrives for a probability distribution represented by $P\{X|S_i\}$, if such a distribution had already been computed earlier and cached, it can be returned immediately. However, usually it will be necessary to send requests to the clique's successors in order to compute the response. Since it can be easily shown that

$$P\{X|S_i\} = \sum_{R_{C_i}} \prod_j P(X \cap T(C_{ij})|S_{ij}) P(R_{C_i}|S_i) \quad (4)$$

where $R_{C_i}$ is the residual nodes of $C_i$, $C_{ij}$ is the $j-th$ child clique of $C_i$, and $S_{ij}$ is the separators between $C_i$ and $C_{ij}$, the request to each child $C_{ij}$ will be

$$P(X \cap T(C_{ij})|S_{ij}). \quad (5)$$

The recursive process continues until it reaches the leaf node or the request can be answered from the cached results.

To handle the evidence, just substitute the observed values into all the clique distributions involving the observed node. This operation is very simple in which the particular dimension of the observed value is simply eliminated and the rest of the distribution remain the same. The substitution needs to be done for all the distributions including the cached results stored in each clique which involves the observed node. After the substitution, all the other operations can be applied on distributions with the reduced dimensions.

As in the SPI algorithm, three major operations are needed in the SEPI algorithm: multiplication, summation and substitution. Multiplication calculates the product of two distributions, summation calculates the sum of a distribution over a set of variables, and substitution calculates the result of substituting an observed value for a node into a distribution.

## 4  Examples

With the network given in Figure 1, we will now illustrate the SEPI algorithm with several query examples. First, assuming the query we are interested is $P(AXS)$, the recursive algorithm works as follows.

- The query $P(AXS)$ is received at the root clique $(AT)$, based on eqn. (2) and (3), a new query $P(XS|T)$ is generated and sent to the successor clique $(TLE)$

- The query $P(XS|T)$ is received at the clique $(TLE)$; similarly, a new query $P(XS|LE)$ is generated and sent to the successor clique $(LEB)$

- The query $P(XS|LE)$ is received at the clique $(LEB)$, based on (5), new queries $P(S|BL)$ and $P(X|EB)$ are generated and sent to the successors $(BLS)$ and $(EBD)$ respectively.

- The query $P(S|BL)$ is received at the clique $(BLS)$ which is available in the cache due to the pre-processing.
- The query $P(X|EB)$ is received at clique $(EBD)$, a new query $P(X|E)$ is generated and sent to the successor clique $(EX)$
- The query $P(X|E)$ is received at clique $(EX)$ which is available.

- At clique $(EBD)$, compute the query $P(X|EB)$ by

$$P(X|EB) = \sum_D P(X|E)P(D|BE) \quad (6)$$

- At clique $(LEB)$, compute the query $P(XS|LE)$ by

$$P(XS|LE) = \sum_B P(S|BL)P(X|EB)P(B|LE) \quad (7)$$

- At clique $(TLE)$, compute the query $P(XS|T)$ by

$$P(XS|T) = \sum_{LE} P(XS|LE)P(LE|T) \quad (8)$$

- At root clique $(AT)$, compute the query $P(AXS)$ by

$$P(AXS) = \sum_T P(XS|T)P(AT) \quad (9)$$

Assume in the second example that the node $E$ is observed and the observed value is $E^*$. To calculate the posterior probability of the same query, we first substitute the observed value into all the distributions in the cliques related to the observed node. These include $P(D|BE)$, $P(B|LE)$, and $P(LE|T)$ in the cliques $(TLE)$, $(LEB)$, and $(EBD)$ respectively. The substitution operation simply eliminates the particular dimension corresponding to the observed value in the distributions. Then the same procedure as described above to calculate the query can be applied using the distributions with new reduced dimensions. The result is therefore,

$$P(AXS|E = E^*) = \sum_T \sum_L [\sum_B [P(S|BL)\sum_D [p(X|E^*)P(D|BE^*)] P(B|LE^*)] P(LE^*|T)] P(AT) \quad (10)$$

## 5   Conclusion

SPI algorithm [1, 2] is the latest inference algorithms in which the emphasis is on the efficient generic query. The main goal of these algorithms is to respond to arbitrary queries in an efficient manner. In these algorithms the network is first converted into a search tree and the probabilities are manipulated by symbolically decomposing or factoring the formulae. These methods are incremental with respect to queries and evidence and have good potential for parallel processing.

In this paper, we develop a similar query algorithm based on the combination of evidence potential algorithm and the SPI inference mechanism. Rather than converting the network into a SPI search tree, we construct a "clique tree" based on the evidence potential algorithm. Additionally, the evidence potential algorithm is used as the pre-processing step where all the necessary probability distributions for answering the query are computed and stored in each clique.

Similar to the SPI algorithm, queries are directed to the root clique of the tree. They are decomposed into queries for the clique's subtrees. This recursive procedure continues until a particular query can be answered. The answer is then computed and returned to the next higher level. The algorithm and the computation are simple. With a similar mechanism for caching and incorporating evidence as in the SPI algorithm, the calculation is also incremental with respect to both query and evidence. However, the SEPI algorithm is more efficient since all the necessary probability distributions are stored in the pre-processing step. A version of the SEPI algorithm as well as the SPI algorithm have been implemented, preliminary results from several examples show that with the prep-processing step, the query process of the SEPI algorithm is faster than the SPI algorithm.

## References

[1] B. D'Ambrosio. Symbolic probabilistic inference in belief nets. 1990.

[2] R. Shachter A. Del Favero and B. D'Ambrosio. Symbolic probabilistic inference: A probabilistic perspective. *Proceeding of AAAI*, 1990.

[3] S. L. Lauritzen and D. J. Spiegelhalter. Local computations with probabilities on graphical structures and their application in expert systems. *Journal Royal Statistical Society B*, 50, 1988.

[4] Judea Pearl. Fusion, propagation, and structuring in belief networks. *Artificial Intelligence*, 29, 1986.

[5] Judea Pearl. *Probabilistic Reasoning in Intelligent Systems: Networks of Plausible Inference*. Morgan Kaufmann Publishers, 1988.

[6] Ross D. Shachter. Intelligent probabilistic inference. In L.N. Kanal and J.F. Lemmer, editors, *Uncertainty in Artificial Intelligence*. Amsterdam: North-Holland, 1986.